\documentclass[10pt,twocolumn,letterpaper]{article}

\usepackage{cvpr}
\usepackage{times}
\usepackage{epsfig}
\usepackage{graphicx}
\usepackage{amsmath}
\usepackage{amssymb}
\usepackage{multirow}
\usepackage{subfigure}
\usepackage{color}

\usepackage[pagebackref=true,breaklinks=true,letterpaper=true,colorlinks,bookmarks=false]{hyperref}

 \cvprfinalcopy 

\def\cvprPaperID{****} 

\ifcvprfinal\fi
\begin{document}

\title{Multi-bin Trainable Linear Unit for Fast Image Restoration Networks}
\author{Shuhang Gu$^1$, Radu Timofte$^1$, Luc Van Gool$^{1,2}$
\\$^1$ ETH, Zurich, $^2$ KU Leuven\\
\{shuhang.gu, radu.timofte, vangool@vision.ee.ethz.ch\}}

\maketitle

\begin{abstract}
Tremendous advances in image restoration tasks such as denoising and super-resolution have been achieved using neural networks. Such approaches generally employ very deep architectures, large number of parameters, large receptive fields and high nonlinear modeling capacity. In order to obtain efficient and fast image restoration networks one should improve upon the above mentioned requirements.

In this paper we propose a novel activation function, the multi-bin trainable linear unit (MTLU), for increasing the nonlinear modeling capacity together with lighter and shallower networks.
We validate the proposed fast image restoration networks for image denoising (FDnet) and super-resolution (FSRnet) on standard benchmarks. We achieve large improvements in both memory and runtime over current state-of-the-art for comparable or better PSNR accuracies.
\end{abstract}

\section{Introduction}

Image restoration refers to the task of estimating the latent clean image from its degraded observation, it is a classical and fundamental problem in the area of signal processing and computer vision.
Recently, deep neural networks (DNNs) have been shown to deliver
standout performance on a wide variety of image restoration tasks.
With massive training data, very deep models have been trained for pushing the state-of-the-art of different restoration tasks.

The idea of DNN-based image restoration method is straight forward: training a neural network to capture the mapping function between the degraded images and their corresponding high quality images.
By stacking the standard convolution layers and ReLU activation functions, the VDSR approach~\cite{VDSR} and the DnCNN approach~\cite{DnCNN} have achieved state-of-the-art performance on the image super-resolution (SR) and image denoising tasks, respectively.
Furthermore, it is perhaps unsurprising that we are able to further improve the results by these models by further increase their layer numbers; since deeper network structure not only have stronger nonlinear modeling capacity but also incorporate input pixels from a larger area.
However, as there are many practical situations where only limited computational resources are available, how to design an appropriate network structure for such conditions is an important research direction.
\begin{figure}[t!]
\label{fig:1}
\centering
\subfigure{
\begin{minipage}[t]{1\linewidth}
\centering
\includegraphics[width=1\textwidth]{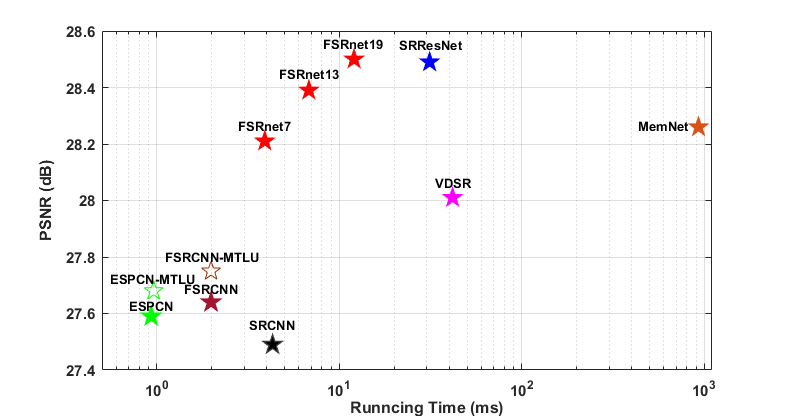}
\end{minipage}
}
\vspace{-0.1cm}
\caption{ Average PSNR on Set 14 ($\times 4$) \textit{\textbf{vs.}} runtime (on Titan X Pascal GPU with Caffe \cite{caffe} toolbox) for processing a $512\times512$ HR image by different approaches. MTLU is able to improve previous fast SR methods without increasing the running time. The proposed FSRnets achieved a good trade-off between speed and performance than state-of-the-art SR methods.} 
\label{fig:sr_teaser}
\vspace{-0.2cm}
\end{figure}

Generally, the key challenge in learning fast image restoration networks is twofold: (i) to incorporate sufficient receptive fields, and (ii) to economically increase the nonlinear modeling capacity of networks.
In order to increase the receptive field of the proposed network, we utilize a similar strategy with recent works~\cite{ESPCN,FFDNET} and conduct all the computations in a lower spatial resolution.
While for the purpose of stronger nonlinear capacity, we propose a multi-bin trainable linear unit (MTLU) as the activation functions used in our networks.
MTLU divides the activation space into equidistant bins and approximates the activation function in each bin with a linear function.
Compared with other parameterization schemes, e.g. summation of different kernel functions~\cite{TNRD,TAF,APL}, the computational burden of MTLU is independent with the accuracy of parameterization.
Such a feat enables us to greatly enhancing the nonlinear capacity of each activation function without significantly increase its computational burden, and achieve a more economic solution for fast image restoration.

We evaluated the proposed MTLU on the classical image super-resolution and image denoising problems.
In Figure~\ref{fig:sr_teaser}, we present the trade-off between speed and performance achieved by different SR approaches. The average PSNR index achieved by different methods for SR the \textit{Set 14 } dataset with scaling factor 4 is reported in relation to the running time required for processing a $512\times512$ high-resolution  image.
By replacing the activation functions used in previous fast SR approaches \cite{ESPCN,FSRCNN} by MTLU, we are able to improve their SR performance without increasing the running time.
Furthermore, the proposed FSRnet is 0.2 up to 0.5dB better in PSNR terms than the benchmark method VDSR~\cite{VDSR} while being 10 to $4\times$ faster, and achieves comparable PSNR with the state-of-the-art SRResNet~\cite{SRResNet} but $3\times$ faster.
For the image denoising task, the proposed method achieved comparable performance with the state-of-the-art approach DnCNN~\cite{DnCNN} but $15\times$ faster.

\section{Related Works}
\subsection{DNN for Image Restoration}
Due to its unparalleled nonlinear modeling capacity, deep neural networks (DNNs) have been widely applied to different image restoration/enhancement tasks~\cite{SRCNN,VDSR,SRResNet,ESPCN,DnCNN,FFDNET,IRCNN}. 
In this part, we only provide a very brief review of DNN-based image denoising and SR approaches, which are the two tasks investigated in this paper.

To deal with SR problem, DNN-based approaches proposed to train a neural network for buliding the mapping function between low-resolution (LR) and high-resolution (HR) images.
Dong~\etal~\cite{SRCNN} firstly proposed a deep learning based model for SR -- SRCNN, a 3 layers CNN with large kernels.
SRCNN~\cite{SRCNN} achieved comparable performance with then state-of-the-art conventional SR approaches based on sparse representations~\cite{SCSR,CSC_SR} and anchored neighborhoods~\cite{A+}, triggering the tremendous investigation of DNN-based SR approaches.
Kim~\etal~\cite{VDSR} proposed VDSR which utilizes a deeper neural network to estimate the residual map between HR and LR image, and achieved superior SR performance.
Recently, Ledig~\etal~\cite{SRResNet} built SRResNet, a very deep neural network of residual blocks, obtaining state-of-the-art SR performance. 
Besides designing deeper networks for pursuing better SR performance, other interesting research directions include investigating of better losses \cite{PerceptualLoss,SRResNet} for generating perceptual plausible SR results, extending the generalization capacity of SR network for different kernels settings and faster SR networks~\cite{ESPCN}.

The application of discriminatively learned networks ~\cite{TNRD,Denoise1,Denoise2,MLP} for image denoising task is earlier than its application on SR tasks.
Different types of networks, including auto-encoder~\cite{Denoise2}, multi-layer perceptron~\cite{MLP} and unfolded inference process of optimization models \cite{ShrinkageField,TNRD} have been suggested for dealing with the image denoising problem.
Recently, Zhang~\etal~\cite{DnCNN} combined recent advances in DNN design and proposed a DnCNN model reaching state-of-the-art performance on denoising tasks with different noise levels.
\subsection{Activation Functions of DNN}
The nonlinear capacity of deep neural networks come from the non-linear activation functions (AFs).
In the study of early years, the designing of activation functions often with strong biological or probability theory motivations, the sigmoid and tanh functions have been suggested for introducing nonlinearity in networks.
While, the recent study of activation functions take more consideration on practical training performance, the Rectified Linear Unit (ReLU) \cite{Relu} function became the most popular AF since it enables better training of deeper networks~\cite{JustifyRelu}.
Different AFs, including the Leaky ReLu (LReLU)~\cite{LeakyRelu}, Exponential Linear Units (ELU)~\cite{ELU} and the Max-Out Unit (MU) \cite{MaxOUT} have been designed for improving the  performance of DNN.

Theoretically, a sufficiently large networks with any of the above hand-crafted AFs can approximate arbitrarily complex functions~\cite{cho2010large}.
However, for many practical applications where the computational resources are limited, the choice of AF affects the capacity of networks greatly.
To improve the model fitting ability of network, He~\etal~\cite{PRelu} extend the original ReLU and propose Parametric ReLU (PReLU) by learning parameters to control the slopes in the negative part for each channel of feature maps.
Besides PReLU~\cite{PRelu}, there are still some more complex parameterization of nonlinear functions~\cite{APL,TAF,TNRD}.
However, these approaches~\cite{APL,TAF,TNRD} share a similar idea of adaptively sum several simple functions (kernel function) to achieve a more complex model and, therefore, their computational burden largely increase with the demand on parameterization accuracy.
Furthermore, some of these functions~\cite{APL,TAF} were designed for classification tasks with fixed input size, the AFs were learned to be spatial variant, making them not suitable for spatial invariant image restoration tasks.
\begin{figure*}[t]
\centering
\subfigure{
\begin{minipage}[t]{0.52\linewidth}
\centering
\includegraphics[width=1\textwidth]{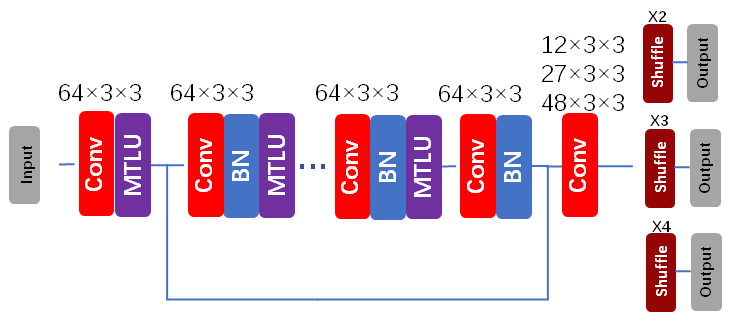}
{\footnotesize  (a) FSRnet (factor 2, 3 and 4)}
\end{minipage}
~
\begin{minipage}[t]{0.44\linewidth}
\centering
\includegraphics[width=1\textwidth]{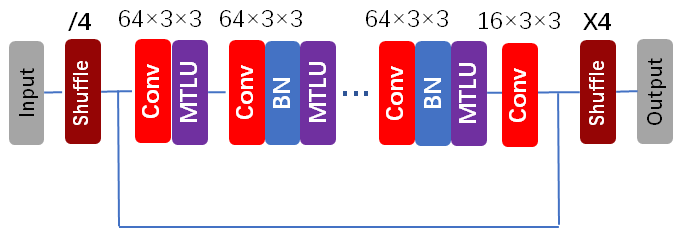}
{\footnotesize  (b) FDnet}
\end{minipage}
}
\caption{The network structures of the proposed FSRnet and FDnet. The hyper parameters for the shuffle layer and convolution layer are shown in the figures. Convolution block with $64\times3\times3$ represents convolution layer with kernel size 3 and 64 output feature maps. Shuffle blocks with parameter $/4$, $\times4$ utilize shuffle operations to enlarge or reduce spatial resolution of input with factor 4.   } 
\label{fig:NNstructure}
\vspace{-0.2cm}
\end{figure*}
\section{Fast Image Restoration with MTLU}
In this section, we introduce the proposed MTLU activation function and the network structures for image super-resolution (FSRnet) and denoising (FDnet).
\subsection{MTLU - Multi-bin Trainable Linear Unit}
The nonlinearity of neural networks comes from the nonlinear activation functions, by stacking some simple operations, \eg convolution and ReLU, the whole networks are able to modeling any nonlinear functions.
However, in many real applications the computational resources are limited and, thus, we are not able to deploy very deep models for capturing the nonlinear function.
This motivated us to improve the capacity of activation functions for better nonlinearity modeling.

Instead of designing fixed activation functions, we proposed to parameterize the activation functions and learn optimal functions for different stage of networks.
We present the following MTLU activation functions, which simply divide the activation space into multiple equidistant bins, and use different linear functions to generate activations in different bins:
\begin{align}
\label{eq:1}
f(x)=
\begin{cases}
a_0x+ b_0,   ~~~~&if ~x\leq c_0;\\
a_kx+ b_k,   ~~~~&if ~c_{k-1}<x\leq c_k;\\
~~~~...\\
a_Kx+ b_K,   ~~~~&if ~c_{K-1}<x.\\
\end{cases}
\vspace{-3mm}
\end{align}
The values $\{c_k\}_{k=0,\dots,K-1}$ are $K$ hyper-parameters for the activation function, and the values $\{a_k, b_k\}_{k=0,\dots,K}$ are parameters to be learned in the training process.
Since the anchor points $c_k$ in our model are uniformly assigned, they are defined by the number of bins (K) and the bin width.
Furthermore,
given the input value $x$, a simple dividing and flooring function can be utilized to find its corresponding bin-index.
Having the bin-index, the activation output can be achieved by an extra multiplication and addition function.

PReLU~\cite{PRelu} activation function can be seen as a special case of the proposed parameterization in which the input space is divided into two bins $(-\infty,0]$ and $(0,\infty)$ and only the parameter $a_0$ is learned, the other parameters $b_0$, $a_1$ and $b_1$ being fixed to 0, 1 and 0, respectively.
The proposed MTLU trains more parameters and is expected to have a stronger nonlinear capacity than PReLU. 
The detailed settings for the bin number as well as bin width will be discussed in the experimental section~\ref{ssc:TrainMTLU}. 
\subsection{Network Structure for Super Resolution (SR)}
\label{ssc:FSRNet}

To tackle SR a CNN is trained to extract local structure from LR images for estimating the lost high frequency details of HR images.
By stacking the simple convolution (CONV) + ReLU operations, VDSR~\cite{VDSR} was shown to deliver very good SR performance.
To obtain a fast SR network, we modify the network structure to (i) process the image in a lower spatial resolution and (ii) use MTLU for improved nonlinear modeling capacity.

Conducting the major SR operations in the LR space was originally adopted in the CSC-SR approach~\cite{CSC_SR}, where the convolution sparse coding is used to decompose the LR input image to get feature maps for SR.
For DNN-based SR approaches, Shi~\etal~\cite{ESPCN} firstly suggested to use the LR image (instead of interpolated image) as input and conduct SR operation in the LR space, and such a strategy has been widely applied in recent DNN-based SR methods \cite{SRResNet,EDSR}.
We follow the same strategy and directly use LR image as input.
Different from recent approach~\cite{SRResNet}, which utilizes larger channel numbers and filter sizes to gradually generate the final HR reconstruction from LR feature maps, we utilize the 64 LR feature maps to generate the shuffled HR image with only one $3\times 3$ convolution layer for the purpose of efficiency.
As we will show in section~\ref{ssc:CompareAF}, MTLU greatly improves the nonlinear capacity of network, enabling us to achieve good SR results with a lower depth (number of layers).
Figure~\ref{fig:NNstructure}(a) illustrates the proposed SR network, FSRNet. 
Since the purpose of this paper is to find a good trade-off between restoration performance and processing speed, and 
 FSRNet is capable to achieve top SR results with a few layers, we did not employ residual blocks in the middle of our networks, and only set one residual connection between the feature maps of first layer and last layer.
\subsection{Network Structure for Image Denoising}
\label{ssc:FDNet}
Different from the SR task, the input and the target noise-free image in the denoising task are of same size.
For the purpose of efficiency, we shuffle the input image and conduct the denoising operations at a lower spatial resolution.
Although processing the shuffled LR multi-channel image helps to reduce the computational burden in the training and testing phases as well as greatly improves the perception field of networks, 
it also has a higher demand on the nonlinear power of each layer since the shuffling operation narrows the network width.
To balance speed and performance, we shuffle the input noisy image with factor 4, \eg a noisy image with size $H\times W\times C$ is shuffled to $H/4\times W/4\times 16C$ as the input to the network.
A similar shuffling strategy has been utilized in a very recent paper~\cite{FFDNET}, however, as \cite{FFDNET} utilized the simple ReLU function, shuffling factor 2 was adopted as a trade-off between speed and performance.
While, with the help of MTLU, the proposed FDnet is able to get comparable results with much less running times.
An illustration of the proposed FDnet structure can be found in Figure~\ref{fig:NNstructure}(b).
\subsection{Discussion}
\noindent
\textbf{MTLU vs. Parametrization}
We emphasize that we choose the formulation of MTLU in~\eqref{eq:1} based on efficiency.
Although several other non-linear approximation approaches have been suggested in different areas of computer vision, we will show in the experimental section that the proposed MTLU is able to deliver good results with lower computational burden.
Concretely, since most of previous approaches~\cite{APL,TAF,TNRD} adopted a summation strategy, the increasing of parameterization accuracy will greatly increase the computational burden in the inference phase.
Recently, Sun~\etal proposed a piece-wise linear function (PLF)~\cite{PLF} for compressive sensing, which uses a group of anchor points to decide the function values between the anchor points.
However, since PLF uses anchors to parameterize the linear functions in the intervals, each anchor affects the linear functions in two adjacent bins.
Consequently, the parameterizations in different intervals affect each other and limit the flexibility of PLF.
Furthermore, PLF is not as efficiency as the proposed MTLU, as we will show in section~\ref{ssc:CompareAPLandPLF}, a 6 layers network with PLF takes $10\%$ more running time than the same network with MTLU.
%
%

\vspace{2mm}
\noindent
\textbf{MTLU vs. Discontinuity}
Another thing which is worth discussion is the discontinuity issue. 
Some readers may raise concerns on the training stability of MTLU due to its discontinuity.
Actually, some recent advances in the field of network compression and DNN-based image compression~\cite{QuantizedNN,BinarizedNN,compression1,compression2,compression3} have shown that the networks work well even with non-continuous functions.
Furthermore, we experimentally found that the training of MTLU is very stable, even without a BN layer to normalize the range of inputs, it still able to deliver good restoration performance.
\section{Experimental Results}
\label{sec:experimental_results}
In this section, we provide experimental results to show the advantage of the proposed models.
First, we discuss some training aspects of the proposed MTLU and conduct experiments to compare MTLU with other activation functions which have been widely used in other image restoration networks.
Then, we compare the proposed networks with representative state-of-the-art SR and denoising networks.

All the experiments are conducted on a computer with Intel Xeon CPU e5-2620, 64 GB of RAM and Nvidia Titan X Pascal GPU.
We evaluate the running time of different networks with Caffe toolbox~\cite{caffe}.
\subsection{Training with MTLU}
\label{ssc:TrainMTLU}
The proposed MTLU has several hyper parameters, in this section, we discuss some training details as well as parameter settings for the MTLU layers used in this paper.

\vspace{2mm}
\noindent
\textbf{MTLU: gradients}
The parameters for the $k-$th bin will be affected by all the signals drop into ($c_{k-1}, c_k$], thus, the gradients with respect to $a_k$ and $b_k$ can be written as follows
\begin{align}
\label{eq:2}
\frac{\partial loss}{\partial a_k}&=\frac{1}{N_k}\sum_{x_i\in S_k} x_i\partial f_i,\\
\frac{\partial loss}{\partial b_k}&=\frac{1}{N_k}\sum_{x_i\in S_k} \partial f_i,
\end{align}
where $N_k$ is the normalization factor that counts the number of signals in each bin, $S_k$ indicate the range corresponding to $a_k$ and $b_k$, $\partial f_i$ denotes the gradient coming from the next layer in position $i$.
In our implementation, we did not count the number of signals laying in each bin, and simply use the bin-number/signal-number as the normalization factor.
The gradient of MTLU parameters is relatively small and we found weight decay will affect the training of MTLU. 
For all the experiments in this paper, we do not conduct weight decay on MTLU.
While, the other parameters are regularized by a weight decay factor of $10^{-4}$ which is the same as \cite{VDSR,DnCNN}.
%

\vspace{3mm}
\noindent
\textbf{MTLU: initialization}
In our experiments, we initialize MTLU as a ReLU function. 
With other initializations, such as random initialization of $\{a_k, b_k\}_{k=0,\dots,K}$ and initialization MTLU as a identity mapping function $f(x)=x$, MTLU is still trainable.
However, we experimentally found that the ReLU initialization often delivers a better  convergence (about 0.05dB for SR \textit{Set 14} with factor 4) than the models trained with
random initialization or identity initialization.
%
%
\begin{figure*}[t!]
\centering
\subfigure{
\begin{minipage}[t]{0.2\linewidth}
\centering
\includegraphics[width=1\textwidth]{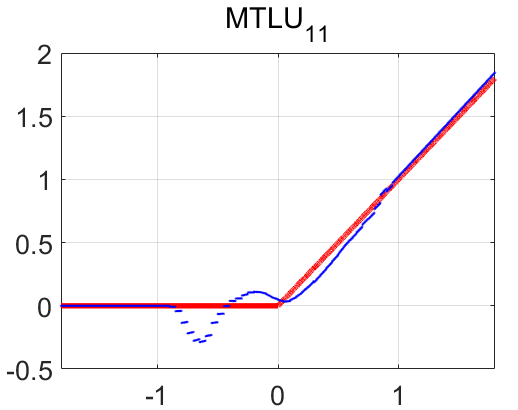}
\includegraphics[width=1\textwidth]{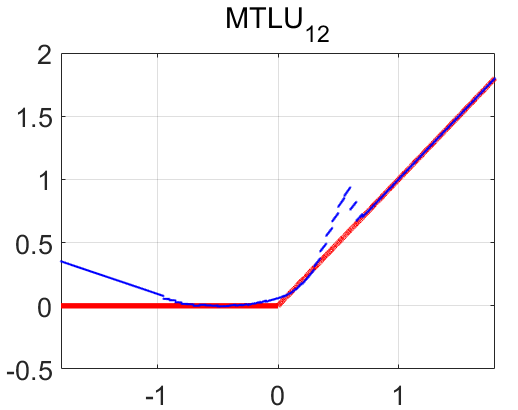}
\end{minipage}
\begin{minipage}[t]{0.2\linewidth}
\centering
\includegraphics[width=1\textwidth]{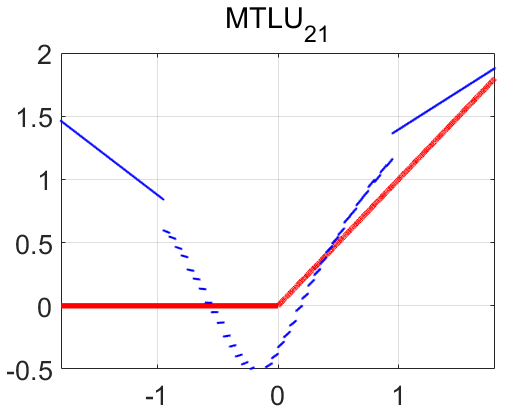}
\includegraphics[width=1\textwidth]{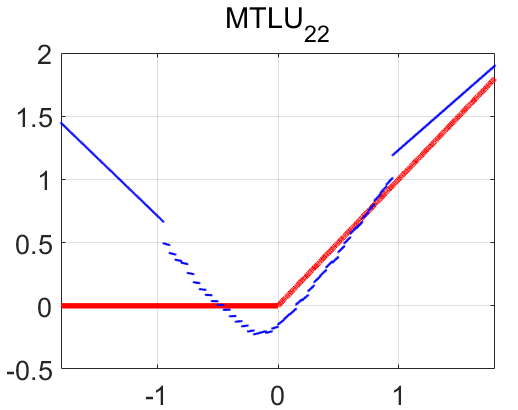}
\end{minipage}
\begin{minipage}[t]{0.2\linewidth}
\centering
\includegraphics[width=1\textwidth]{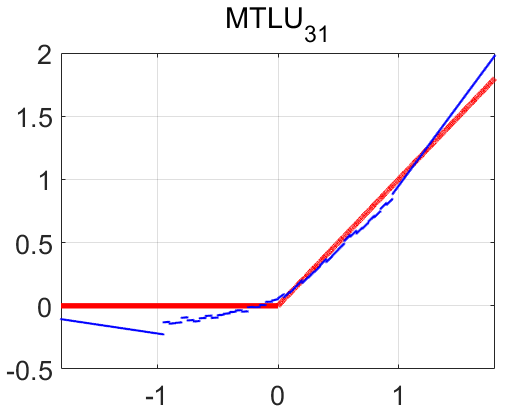}
\includegraphics[width=1\textwidth]{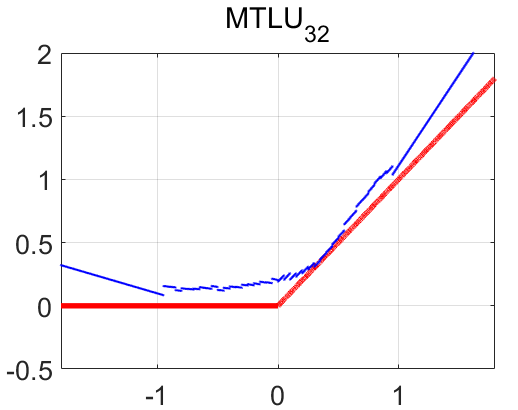}
\end{minipage}
\begin{minipage}[t]{0.2\linewidth}
\centering
\includegraphics[width=1\textwidth]{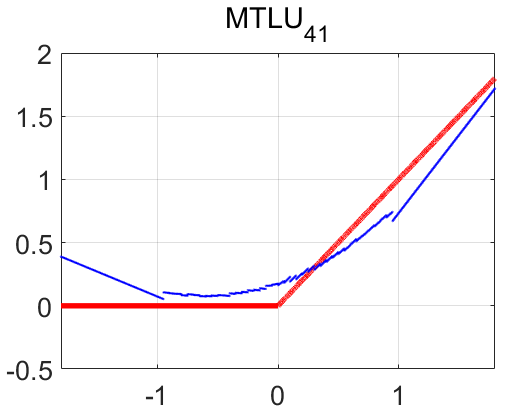}
\includegraphics[width=1\textwidth]{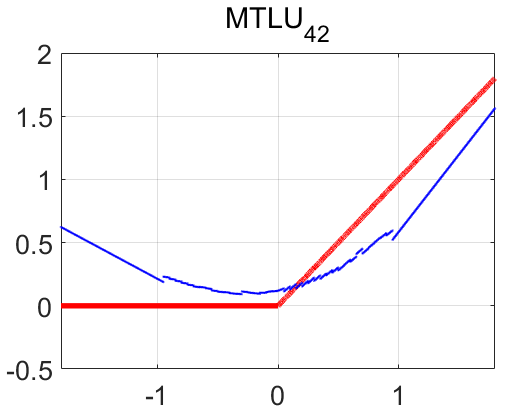}
\end{minipage}
\begin{minipage}[t]{0.2\linewidth}
\centering
\includegraphics[width=1\textwidth]{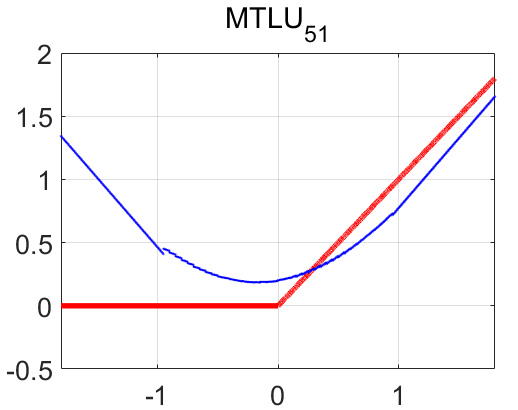}
\includegraphics[width=1\textwidth]{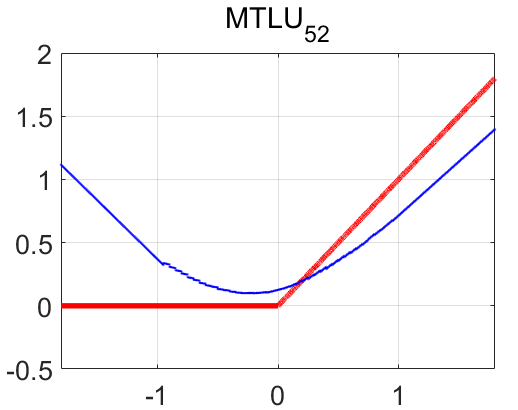}
\end{minipage}
}
\caption{Examples of learned \textcolor{blue}{MTLUs} in FSRnet$_7$. MTLU$_{ij}$ denotes the activation function for the $j$-th channel of layer $i$. ReLU function (in \textcolor{red}{red}) is included for reference.}
\label{fig:MTLUs}
\end{figure*}
\begin{table}[htp]
\centering
\small
\caption{SR Results by FSRnet$_{7}$ with different bin-width on \textit{Set 5} (factor 4). Number of bins are fixed to  2/bin-width.}
\begin{tabular}{c|ccccc}
Bin-width& ~~0.025~~& ~~0.05~~& ~~0.1~~& ~~0.2~~& ~~0.5~~\\
\hline
PSNR [dB]&~31.49~&~31.52~&~31.48~&~31.49~&~31.43~\\
\end{tabular}
\label{tab:binwidth}
\end{table}
\begin{table*}[htp]
\centering
\caption{SR results on Set 5 ($\times 4$) and run time for processing a $512\!\times\!512$ image by FSRnet$_7$-MTLU and variations of FSRnet$_7$-APL \cite{APL}.}
{
\begin{tabular}{c|c|c|c|c|c|c}
AFs&MTLU$_{40}$&$APL_2$&$APL_5$&$APL_{10}$&$APL_{20}$&$APL_{40}$\\
\hline
PSNR[dB]/Run Time[ms]&31.52/4.0&31.31/4.7&31.34/7.8&31.35/16.0&31.29/34.1&31.32/87.6\\
\end{tabular}
}
\label{tab:APL}
\end{table*}

\vspace{3mm}
\noindent
\textbf{MTLU: number of bins and bin width}
The number of bins as well as bin-width determines the parameterization accuracy and range of the proposed MTLU layer.
In both FDnet and FSRnet proposed structures, we have a Batch Normalization (BN) layer to help us adjust the range of inputs.
We thus only carefully parameterize the activations between the range of [-1, 1], since most of the inputs of MTLU lies in this range.
Note that for input signals out of the range [-1, 1], MTLU still generates valid activations, but for all the values $x \leq c_0$ or $x>c_{N-1}$, they share the same two groups of parameters $\{a_0, b_0\}$ or $\{a_N, b_N\}$.

\begin{table*}[htp!]
\centering
\caption{SR results on Set 5 ($\times 4$) and run time for processing a $512\times512$ image by variations of FSRnet$_7$-MTLU and FSRnet$_7$-PLF \cite{PLF}.}
\begin{tabular}{c|c|c|c|c|c|c}
AFs&MTLU$_{20}$&MTLU$_{40}$&MTLU$_{80}$&PLF$_{20}$&PLF$_{40}$&PLF$_{80}$\\
\hline
PSNR/Run Time&31.50/4.0&31.52/4.0&31.54/4.0&31.34/4.3&31.39/4.4&31.49/4.4\\
\end{tabular}
\label{tab:PLF}
\end{table*}

After fixing the parameterization ranges, we only need to choose the bin-width and the number of bins is obtained as 2/bin-width.
To choose the bin-width values, we train a group of 7 layers FSRnet (FSRnet$_7$) with different bin-width values, and evaluate the SR results by different models.
The average PSNR values by different settings for SR the \textit{Set 5} data set with zooming factor 4 are shown in Table~\ref{tab:binwidth}.
Intuitively, a smaller bin-width enables a higher parameterization accuracy of MTLU, and is expected to improve the nonlinearity modeling capacity of the network.
However, this comes at the price of introducing more parameters (larger number of bins) to cover the range between (-1,1).
Furthermore, as shown in Table~\ref{tab:binwidth}, the result obtained with a bin-width smaller than 0.025 is worse than the ones with larger bin-widths. A possible reason is that the number of signals in certain bins may be very small rendering the training process unstable.
In our experiments, we divided the space between $-1$ and $1$ into 40 bins (bin-width 0.05) and learn different AFs for different channels. 
Thus, for a network with 64 channels of feature map, the number of parameters of each activation function is $64\times80=5120$, which is less than $1/7$ of the parameter number of one $3\times3$ convolution layer.
The AFs of the first and second feature maps in each MTLU layers of learned FSRnet$_7$ are shown in Figure~\ref{fig:MTLUs}.
\subsection{Comparison with other non-linear parameterization approaches}
\label{ssc:CompareAPLandPLF}
In this part, we compare MTLU with two representative non-linear parameterization approaches, e.g. APL \cite{APL} and PLF \cite{PLF}.
APL \cite{APL} is one representative non-linear parameterization approach, it  use summation of ReLU-like units to increase the capacity of activation function.
As APL $h_i(x)=\max(0,x)+\sum_sa_i^s\max(0,-x+b_i^s)$ was designed for high-level vision tasks and the AF is variant at different position $i$, it is not directly applicable for restoration tasks with different input sizes.
Thus, we remove the spatial variant part and adopt the same activation in different channels of feature map (the same setting as adopted for MTLU).
We adopt APL with different number of kernels in FSRnet$_7$; the avg. PSNR on Set 5 ($\times 4$) and runtime for processing a $512\times512$ image of FSRnet$_7$-APL and the proposed FSRnet$_7$-MTLU (40 bins) are shown below. 
FSRnet$_7$-MTLU is much faster and $\sim$0.2dB better than FSRnet$_7$-APL.

PLF is another recently proposed non-linear parameterization approach, it uses a group of anchor points to determine the function values in the intervals.
To compare the MTLU with PLF \cite{PLF}, we adopt the two AFs in FSRnet$_7$.
Specifically, we keep both the bin-width of MTLU and anchor interval of PLF as 0.05, and use different bin-numbers (anchor numbers).
The avg. PSNR on Set 5 ($\times 4$) and runtime for processing a $512\times512$ image of FSRnet$_7$-PLF and the proposed FSRnet$_7$-MTLU (40 bins) are shown below.
As the table shows, FSRnet$_7$-MTLU achieves better results and is about $10\%$ faster than FSRnet$_7$-PLF \cite{PLF}.
Furthermore, when the PLF adopt relatively small number of anchors, it can not deliver good SR results.
One possible reason is that PLF uses anchors to parameterize the linear functions in the intervals, each anchor will affect the linear functions in two adjacent bins; 
as a result, when the intervals do not cover all the possible value range of input, e.g., PLF20 and PLF40 only covers input range of [-0.5,0.5] and [-1,1], the values outside the range will greatly affect the parameterization and PLF can not achieve performance as good as MTLU, which decouples the functions in each bin.
Such a drawback of PLF may limits its application on feature maps which we do not have prior knowledge on the value ranges (\eg feature maps without BN).
Actually, PLF \cite{PLF} was originally applied on images which has a strict fixed range.

\begin{table*}[htp]
\centering
\caption{Super-resolution PSNR results [dB] on Set 5 and Set 14 ($\times 4$) and running time for processing a $512\times512$ image by variations of ESPCN \cite{ESPCN} and FSRCNN \cite{FSRCNN}.}
\small
{
\begin{tabular}{c|c|c|c|c|c|c}
\hline
\hline
Networks&ESPCN$_{ReLU}$&ESPCN$_{PReLU}$&ESPCN$_{MTLU}$&FSRCNN$_{ReLU}$&FSRCNN$_{PReLU}$&FSRCNN$_{MTLU}$\\
\hline
PSNR on Set 5&30.66&30.67&30.82&30.73&30.76&30.96\\
\hline
PSNR on Set 14&27.60&27.60&27.68&27.61&27.64&27.75\\
\hline
Ave. Runtime (ms)&0.94&0.94&0.95&1.97&1.98&1.98\\
\hline\hline
\end{tabular}
}
\label{tab:espcnandfsrcnn}
\end{table*}
\subsection{Comparison with other activation functions}
\label{ssc:CompareAF}
In this part, we compare MTLU with other AFs recently adopted in image restoration networks~\cite{DnCNN,SRResNet,MOUSR,FFDNET}.

The comparison includes the most commonly utilized ReLU~\cite{Relu} function; the PReLU~\cite{PRelu} function which has been adopted in the state-of-the-art SR algorithm SRResNet~\cite{SRResNet}, and the Max-out Unit (MaxOut)~\cite{MaxOUT} which has been adopted in a recently proposed SR approach~\cite{MOUSR}.
We compare different activation functions on both the image SR and denoising tasks.
Specifically, the fast SR approaches ESPCN \cite{ESPCN},FSRCNN \cite{FSRCNN}, the state-of-the-art SR approach SRResNet \cite{SRResNet} as well as the proposed FSRnet (Fast SR Net, see Section~\ref{ssc:FSRNet}) are utilized to compare different AFs on the SR task;
and 
the state-of-the-art denoising network structure DnCNN~\cite{DnCNN} as well as the proposed FDNet (Fast Denoising Net, see Section~\ref{ssc:FDNet}) are utilized to compare different AFs on the denoising task.

For fast SR approaches: ESPCN \cite{ESPCN} and FSRCNN \cite{FSRCNN}, since the network structure is simple, we just replace the AFs in the two networks to compare different AFs.
While, in order to thoroughly compare the performance of different AFs with more complex network structures: SRResNet \cite{SRResNet}, FSRnet, DnCNN \cite{DnCNN}, FDnet,
 we vary the basic building blocks (residual blocks for SRResNet~\cite{SRResNet} and CONV+BN+AF blocks for the other networks) in the 4 network structures and compare the performance/speed curves of networks with different AFs.
The details for the employed settings of the different network structures are described in the next.

\subsubsection{ESPCN and FSRCNN with different AFs}
ESPCN \cite{ESPCN} and FSRCNN \cite{FSRCNN} are two recently proposed fast SR approaches, the network structures were designed to achieve good SR performance with small computational burden.
ESPCN \cite{ESPCN} adopts three convolution+ReLU layers, the kernel sizes of different layers are \{5, 3, 3\} and the feature map numbers of the first and second layers are \{64, 32\}.
We follow the same setting and only conduct SR operation for the illumination channel.
ESPCN \cite{ESPCN} with PReLU \cite{PRelu} and the proposed MTLU are trained to show the advantage of MTLU.
FSRCNN \cite{FSRCNN} is another fast SR network structure which adopts more layers but less feature maps to trade off the SR performance and inference speed.
The original FSRCNN \cite{FSRCNN} utilize PReLU \cite{PRelu} as AF, we provide variations of FSRCNN-ReLU and FSRCNN-MTLU to compare different AFs.

For all the variations of FSRCNN and ESPCN, we adopt the DIV2K \cite{DIV2K} as training set.
The learning rate is initialized as $1^{-3}$ and we divide the learning rate every 50K iterations until the learning rate is less than $1^{-5}$.
We train all the networks with Adam \cite{Adam} solver, the default parameters of Adam solver has been adopted in all the experiments in this paper.
The PSNR results ($\times 4$) on Set 5 and Set 14 by different methods are shown in table \ref{tab:espcnandfsrcnn}, the running time of different methods for processing a $512\times 512$ image are also provided.
One can see in the table that MTLU greatly improves the fast SR approaches and almost does not introduce any extra computational burden.

\subsubsection{SRResNet with different AFs}
SRResNet~\cite{SRResNet} utilizes a large number of residual blocks, conducts most of the computation in the LR space, to then use large kernels and 2 shuffle steps to gradually reconstruct the HR estimation with the LR feature maps.
We change the number of residual blocks to obtain different operating points for SRResNet.
For the proposed MTLU as well as the compared AFs (MaxOut, ReLU and PReLU) we train 4 networks with 4, 8, 12 and 16 residual blocks.

For all the variations of networks, we use DIV2K \cite{DIV2K} as training set.
We follow the original experimental settings in~\cite{SRResNet} and crop $96\times96\times3$ subimages from the training dataset with batch size of 16.
We initialize the learning rate as $1\time 10^{-3}$, for networks with residual blocks 4, 8, 12 and 16, the  learning rate are divided by 2 every 80K, 100K, 120K and 150K iterations.
The training process stops when the learning rate is less than $1\time 10^{-5}$.

\begin{figure}[t]
\centering
\subfigure{
\begin{minipage}[t]{0.5\linewidth}
\centering
\includegraphics[width=1\textwidth]{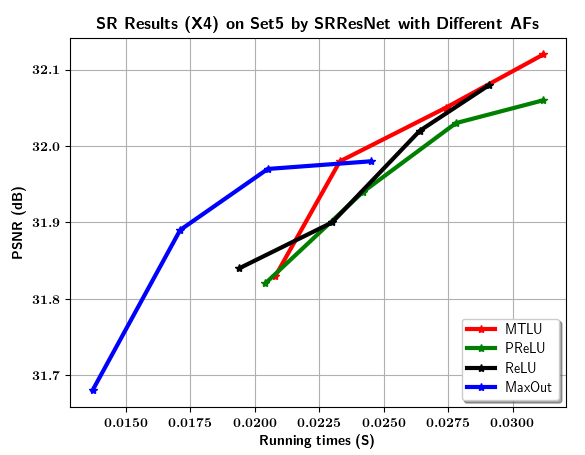}
{\footnotesize  (a)}
\end{minipage}
\begin{minipage}[t]{0.5\linewidth}
\centering
\includegraphics[width=1\textwidth]{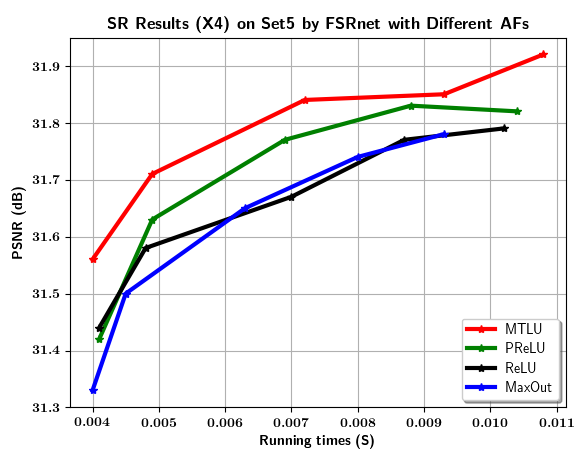}
{\footnotesize  (b)}
\end{minipage}
}
\caption{SR (Set5, zooming factor 4) results by different network structure with different AFs. (a) SR results by SRResNet \cite{SRResNet} with different AFs, the Markers represent SRResNet \cite{SRResNet} variations with residual block numbers 4, 8, 12 and 16, respectively; (b) SR results by FSRnet with different AFs, the Markers represent FSRnet variations with convolution layer numbers 7, 9, 11, 13 and 17, respectively.} 
\label{fig:SRAF}
\end{figure}

The SR (zooming factor 4) results by different settings are shown in Figure \ref{fig:SRAF} (a).
Overall, the MTLU-based networks achieves the best performance among the competing approaches.
By replacing PReLU with the proposed MTLU, we improve the performance of SRResNet from 32.05dB \cite{SRResNet} to 32.12dB on Set 5.

\subsubsection{FSRnet with different AFs}
We further evaluate the effectiveness of MTLU on the proposed FSRnet structure.
We use the same training data as we used for training SRResNet~\cite{SRResNet}, and train FSRnet with 7, 9, 11, 13, and 17 layers, respectively.
The initial learning rate is $1\time 10^{-3}$ and divided by 2 every 80K, 80K, 120K, 120K and 160K for different layer numbers.

The SR results by different settings are shown in Figure~\ref{fig:SRAF} (b).
Compared with other AFs, the MTLU-based network is able to get a better trade off between the processing speed and SR performance.

\subsubsection{DnCNN with different AFs}
DnCNN~\cite{DnCNN} with 17 convolution layers working on the full-resolution input image is a state-of-the-art algorithm for image Gaussian noise removal.
%
To compare different AFs, in DnCNN we replaced ReLU with MaxOut, PReLU and MTLU, resp., and scaled the basic DnCNN structure to 4, 8, 12 and 16 layers by changing the number of CONV+BN+ReLU blocks.
We collected 80,000 $72\times72$ image crops from the 400 training images used by the DnCNN paper~\cite{DnCNN}, and white Gaussian noise with $\sigma=50$ was added into the clean images to generate the noisy input images.
For all the models, we set the batch size as 32.
The learning rate for different models was initialized with $1\times 10^{-3}$, and decayed the learning rate by 2 every 60K, 80K, 100K and 120K iterations for models with layer numbers 4, 8, 12 and 16, respectively.
The training process was stopped when the learning rate dropped below $1\times 10^{-5}$.

\begin{figure}[t]
\centering
\subfigure{
\begin{minipage}[t]{0.5\linewidth}
\centering
\includegraphics[width=1\textwidth]{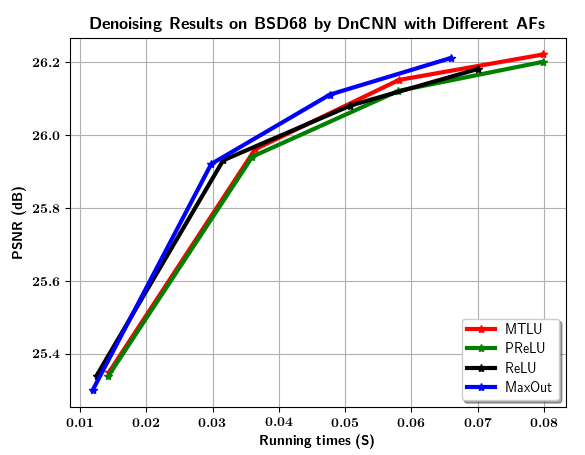}
{\footnotesize  (a)}
\end{minipage}
\begin{minipage}[t]{0.5\linewidth}
\centering
\includegraphics[width=1\textwidth]{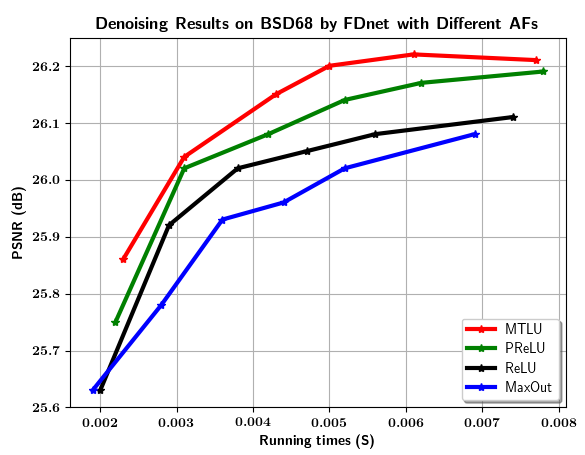}
{\footnotesize  (b)}
\end{minipage}
}
\caption{Denoising results (noise level 50) on the BSD68 dataset by different network settings. (a) Denoising results by DnCNN with different AFs, the Markers represent DnCNN variations with layer numbers 4, 8, 12 and 16, respectively; (b) Denoising results by FDnet with different AFs, the Markers represent FDnet variations with layer numbers 4, 6, 8, 10, 12 and 16, respectively.} 
\label{fig:DenoiseAF}
\end{figure}

The denoising results by different network structures on the BSD68 evaluation set are shown in Figure~\ref{fig:DenoiseAF} (a).
Generally, for the same number of layers, the MTLU achieves the best performance among different AFs.
However, as DnCNN conducts all the operations on the full-resolution images, compared with the nonlinear mapping capacity, the receptive field may be another more important bottleneck to achieve good performance.
With the 64 full-resolution feature maps to provide relatively redundant space for nonlinear modeling, all the 4 AFs are able to deliver high quality denoising results.
Even the MaxOut function, which halves the number of feature maps, achieves comparable denoising results with the other AFs.
In this case, the marginal nonlinear mapping capacity introduced by MTLU can not compensate for its extra computational burden, and the MaxOut \cite{MaxOUT} function is the best choice to get a trade-off between efficiency and performance.

\begin{figure*}[htp]
\centering
\subfigure{
\begin{minipage}[t]{0.195\linewidth}
\centering
\includegraphics[width=1\textwidth]{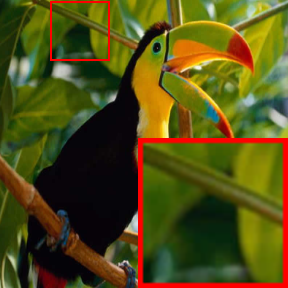}
{\scriptsize  Ground Truth}

{\scriptsize  (PSNR/Run Time)}
\end{minipage}
\begin{minipage}[t]{0.195\linewidth}
\centering
\includegraphics[width=1\textwidth]{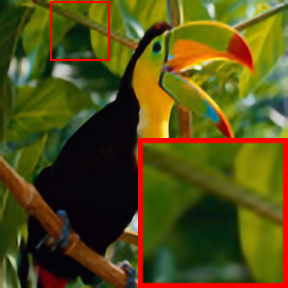}
{\scriptsize  SRCNN \cite{SRCNN}}

{\scriptsize  (35.63 dB/1.9 ms)}
\end{minipage}
\begin{minipage}[t]{0.195\linewidth}
\centering
\includegraphics[width=1\textwidth]{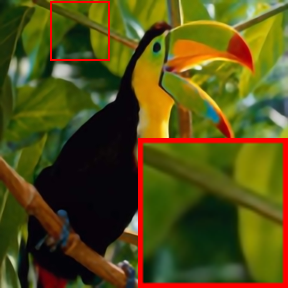}
{\scriptsize VDSR \cite{VDSR}}

{\scriptsize  (36.76 dB/19.3 ms)}
\end{minipage}
\begin{minipage}[t]{0.195\linewidth}
\centering
\includegraphics[width=1\textwidth]{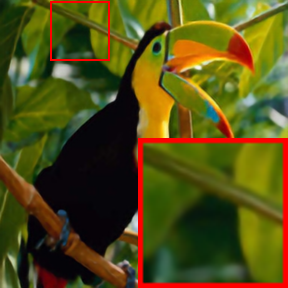}
{\scriptsize FSRnet$_{7}$}

{\scriptsize  (36.94 dB/2.4 ms)}
\end{minipage}
\begin{minipage}[t]{0.195\linewidth}
\centering
\includegraphics[width=1\textwidth]{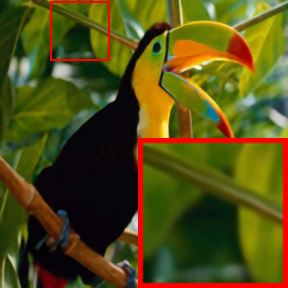}
{\scriptsize FSRnet$_{19}$}

{\scriptsize  (37.88 dB/6.6 ms)}
\end{minipage}
}
\caption{SR results of the \textit{bird} image by different methods (zooming factor 3). The Running times are evaluated on a Titan X Pascal GPU.} 
\label{fig:sr3}
\end{figure*}
\begin{table*}[htp!]
\centering
\caption{Super-resolution PSNR results [dB] by different methods.}
\label{tab:SR_results}
\resizebox{\linewidth}{!}
{
\begin{tabular}{c|c|ccccc|ccc }
\hline\hline
Dataset& Factor& ~~SRCNN~\cite{SRCNN}~~& ~~VDSR~\cite{VDSR}~~& ~LapSRN~\cite{LAPSRN}~&MemNet \cite{MEMNET}&~~SRResNet~\cite{SRResNet}~~& ~~FSRnet$_{7}$~~& ~~FSRnet$_{13}$~~& ~~FSRnet$_{19}$~~\\
\hline\hline
\multirow{ 3}{*}{Set 5}&      $\times2$ &36.66&37.53&37.52&\color{blue}37.78&-&37.58&37.73&\color{red}37.82\\
    &   $\times3$  &32.75&33.66&-&\color{blue}34.09&-&33.71&34.07&\color{red}34.13\\
     &    $\times4$  &30.48&31.35&31.54&31.74&\color{red}32.05&31.52&31.82&\color{blue}31.89\\
      \hline
\multirow{ 3}{*}{Set 14}&  $\times2$ &32.42&33.03&33.08&33.28&-&33.19&\color{blue}33.31&\color{red}33.47\\
    & $\times3$  &29.28&29.77&-&30.00&-&29.93&\color{blue}30.16&\color{red}30.22\\
    & $\times4$ &27.49&28.01&28.19&28.26&\color{blue}28.49&28.22&28.42&\color{red}28.51\\
      \hline
\multirow{ 3}{*}{BSD 100}&   $\times2$ &31.36&31.90&31.80&\color{blue}32.08&-&31.89&32.02&\color{red}32.10\\
    & $\times3$ &28.41&28.82&-&\color{blue}28.96&-&28.80&\color{blue}28.96&\color{red}29.01\\
     &    $\times4$  &26.90&27.29&27.32&27.40&\color{red}27.58&27.31&27.45&\color{blue}27.52\\
\hline\hline
\end{tabular}
}
\end{table*}
\begin{table*}[htp!]
\centering
\caption{Runtimes [ms] by SR methods processing a $512\times512$ pixels HR image.}
\resizebox{\linewidth}{!}
{
\begin{tabular}{c|cccc|ccc}
\hline\hline
 Factor& ~~SRCNN~\cite{SRCNN}~~& ~~VDSR~\cite{VDSR}~~&~MemNet \cite{MEMNET}&~~SRResNet~\cite{SRResNet}~~& ~~FSRnet$_7$~~& ~~FSRnet$_{13}$~~& ~~FSRnet$_{19}$~~\\
\hline
$\times2$& 4.3&42.2&926.8&-&9.7&18.5&27.4\\
$\times3$& 4.2&42.0&930.1&-&6.0&11.1&14.2\\
$\times4$& 4.3&41.7&928.0&31.2&3.9&6.8&8.9\\
\hline\hline
\end{tabular}
}
\label{tab:SR_time}
\end{table*}
\subsubsection{FDnet with different AFs} 
FDnet shuffles the input noisy images with factor $1/4$, the receptive field of network increases quickly with the increase of number of layers.
Furthermore, the same number of 64 feature maps in FDnet need to process the input shuffled image cubic with 16 channels.
Due to both mentioned reasons make FDnet has a higher demand on nonlinear capacity of each layers.

We used the same 80000 image crops and trained a group FDnet with 4, 6, 8, 12, and 16 convolution layers, resp.
For all the models, we set the batch size as 32, and trained them with Adam solver.
The learning rate for different models was initialized with $1\times 10^{-3}$, and divided the learning rate by 2 every 60K, 80K, 100K, 120K and 120K iterations for models with layer numbers of 4, 6, 8, 12 and 16, respectively.
The training process stopped when the learning rate dropped below $1\times 10^{-5}$.

The PSNR indexes by different variations on the BSD68 dataset are shown in Figure~\ref{fig:DenoiseAF} (b).
From the figure, one can easily see that the proposed MTLU achieves much better results than the competing AFs when combined with the proposed FDnet structure.
For equal number of layers, the MTLU-based network outperforms the MaxOut, ReLU and PReLU based networks by about 0.16, 0.12 and 0.08 dB.
Furthermore, since we conduct all the operations at a LR space, the adoption of MTLU introduces only a negligible computation burden in comparison with the compared AFs. The processing speed by FDnet-PRelu and FDnet-MTLU are almost the same.

\subsection{Comparison with state-of-the-art SR algorithms}
In this section, we compare the proposed FSRnet with other SR methods.
The comparison methods include five typical CNN-based SR approaches, \eg the seminal CNN-based SR approach SRCNN~\cite{SRCNN}, the benchmark VDSR method~\cite{VDSR} and current state-of-the-arts LapSRN \cite{LAPSRN}, MemNet \cite{MEMNET} and SRResNet~\cite{SRResNet}.
We compare different methods on zooming factors 2, 3 and 4, and evaluate the SR results on 3 commonly used datasets, \eg Set 5, Set 14 and BSD100, following the settings from~\cite{A+,SRCNN,VDSR,SRResNet}.
The results by the competing methods are provided by the original authors, for the SRResNet~\cite{SRResNet} approach, only the results for zooming factor 4 was provided.

To have a better understanding of the proposed FSRnet, we provide three versions of FSRnet, with 7, 13, and 19 layers namely  FSRnet$_7$, FSRnet$_{13}$, and FSRnet$_{19}$, that trade-off between performance and speed.
All the models of our FSRnet are trained on the DIV2K~\cite{DIV2K} dataset. 
For different zooming factors, we collected training images to make the networks with a spatial resolution of $24\times24$ in the training phase, which means the corresponding HR training images for factors 2, 3, and 4 are with size of 96, 72 and 96, respectively.
We set the batch size as 32, and use the Adam solver to train our models.
For the training of FSRnet$_7$, FSRnet$_{13}$ and FSRnet$_{19}$ models, we initialize the learning rate as 0.001, and divide the learning rate by 2 every 80K, 120K and 200K iterations.
The training process stops when the learning rate is less than $1\times10^{-5}$.

The SR results by different approaches are shown in Table~\ref{tab:SR_results}, and the running time by different models for processing a $512\times512$ image are shown in Table~\ref{tab:SR_time}.
The proposed FSRnet$_7$ achieved slightly better results with VDSR~\cite{VDSR}.
However, the inference of FSRnet$_7$ is much faster ($\times 4$, $\times 7$ and $\times 10$ for zooming factors 2, 3 and 4) than the VDSR \cite{VDSR} approach.
Our FSRnet$_{19}$ achieves PSNR results comparable with the state-of-the-art SRResNet~\cite{SRResNet}, while being more than $3\times$ faster. 
Figure~\ref{fig:sr_teaser} provide a visualization of the trade-off power of our FSRnet model equipped with MTLU.

\begin{figure}[t!]
\centering
\subfigure{
\begin{minipage}[t]{0.5\linewidth}
\centering
\includegraphics[width=1\textwidth]{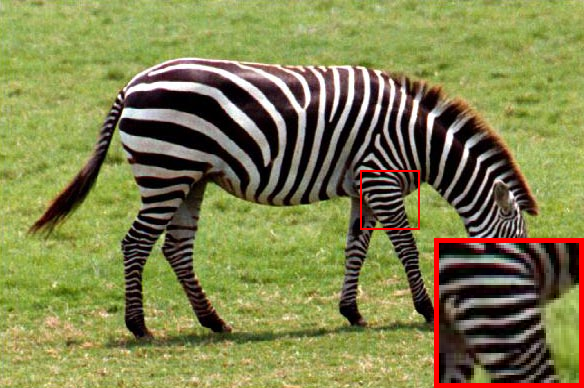}
{\scriptsize  Ground Truth}

{\scriptsize  (PSNR/Run Time)}
\includegraphics[width=1\textwidth]{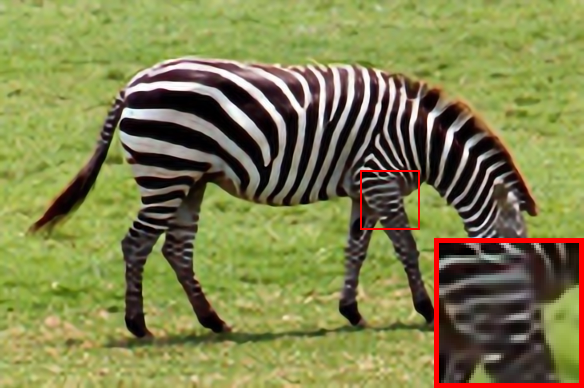}
{\scriptsize  VDSR \cite{VDSR}}

{\scriptsize  (26.94 dB/36.5 ms)}
\includegraphics[width=1\textwidth]{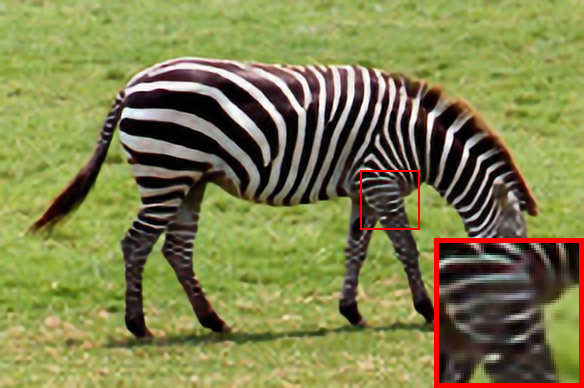}
{\scriptsize  FSRnet$_7$}

{\scriptsize  (26.94 dB/3.4 ms)}
\end{minipage}
\begin{minipage}[t]{0.5\linewidth}
\centering
\includegraphics[width=1\textwidth]{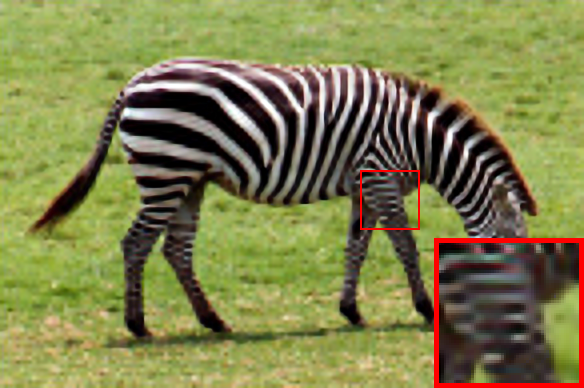}
{\scriptsize  SRCNN \cite{SRCNN}}

{\scriptsize  (24.75 dB/3.8 ms)}
\includegraphics[width=1\textwidth]{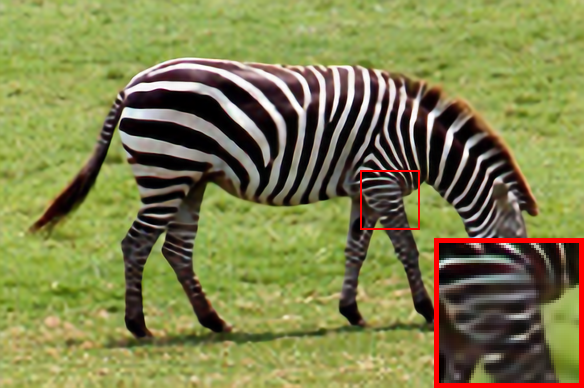}
{\scriptsize  SRResNet \cite{SRResNet}}

{\scriptsize  (27.01 dB/27.2 ms)}
\includegraphics[width=1\textwidth]{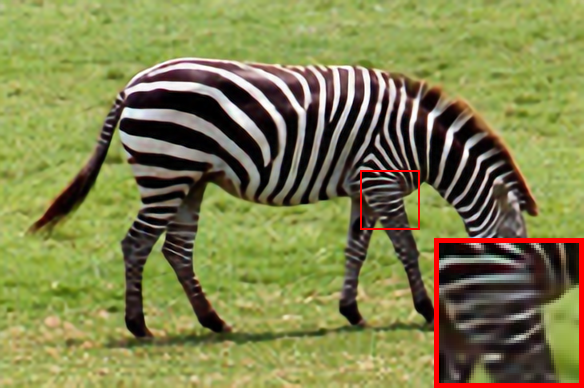}
{\scriptsize  FSRnet$_{18}$}

{\scriptsize  (27.26 dB/7.8 ms)}
\end{minipage}
}
\caption{SR results of the \textit{zebra} image by different methods (zooming factor 4).  The running times are computed on a Titan X Pascal GPU.} 
\label{fig:sr4}
\end{figure}

In Figure \ref{fig:sr3} and \ref{fig:sr4}, we provide some visual examples of the SR results.
One can see that  even the proposed FSRnet$_7$ model can achieve better SR results than the benchmark VDSR \cite{VDSR} approach with about only $1/10$ of the running time.
Compared with  SRResNet \cite{SRResNet}, FSRnet$_{19}$ delivers comparable results with less than $1/3$ of running time.

\subsection{Comparison with state-of-the-art denoising algorithms}
In this section, we compare the proposed FDnet with state-of-the-art denoising algorithms.
Besides the CNN-based approach DnCNN \cite{DnCNN}, several conventional denoising algorithms are utilized as comparison methods, which includes state-of-the-art unsupervised approaches BM3D \cite{BM3D} and WNNM\cite {WNNM_CVPR}, and one discriminative learning approach TNRD \cite{TNRD}.
To achieve a balance between denoising performance and speed, we set the number of convolution layers as 10 in our FDnet.
As shown in Table \ref{tab:ID_time}, our FDnet$_{10}$ is more than 10 times faster than the DnCNN approach in processing a $512\times512$ noisy image.

We used the same 80000 image crops to train our FDnet, the learning rate is initialized as $1\times 10^{-3}$ and divided by 2 every 100K iterations.
The denoising results by different approaches are shown in Table \ref{tab:ID_results}.

\begin{table}[htp!]
\centering
\caption{Runtime [ms] of DnCNN~\cite{DnCNN} in comparison with our FDnet when processing a $512\times512$ noisy image. Both times were computed on a Titan X Pascal GPU.}
\label{tab:ID_time}
\begin{tabular}{c|cc}
\hline\hline
 & ~~DnCNN~\cite{DnCNN}~~& ~~\textbf{FDnet (our)}~~\\
\hline
Runtime [ms]& 74.8&\textbf{5.0}\\
\hline\hline
\end{tabular}
\end{table}

\begin{figure}[t!]
\centering
\subfigure{
\begin{minipage}[t]{0.5\linewidth}
\centering
\includegraphics[width=1\textwidth]{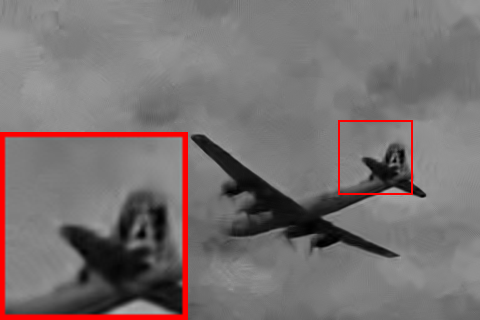}
{\scriptsize  BM3D \cite{BM3D}}
\includegraphics[width=1\textwidth]{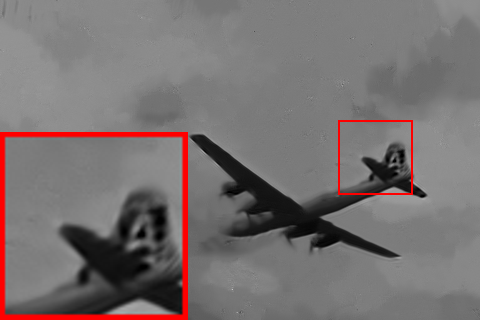}
{\scriptsize  WNNM \cite{WNNM_IJCV}}
\includegraphics[width=1\textwidth]{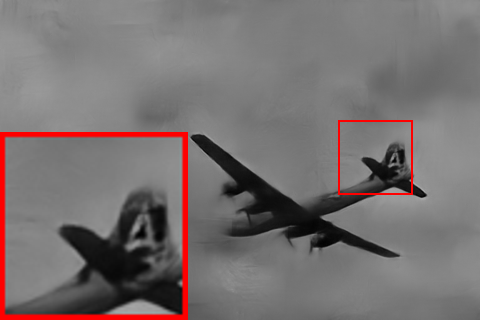}
{\scriptsize  FDnet (\textbf{our})}
\end{minipage}
\begin{minipage}[t]{0.5\linewidth}
\centering
\includegraphics[width=1\textwidth]{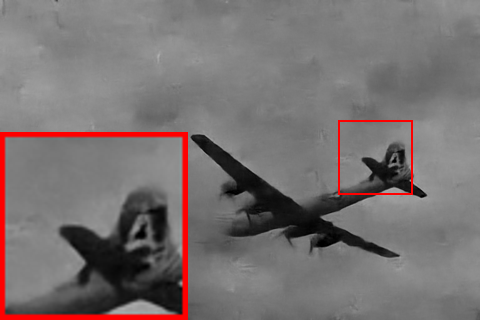}
{\scriptsize  TNRD \cite{TNRD}}
\includegraphics[width=1\textwidth]{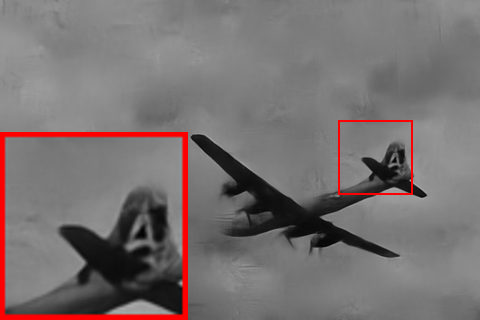}
{\scriptsize  DnCNN \cite{DnCNN}}
\includegraphics[width=1\textwidth]{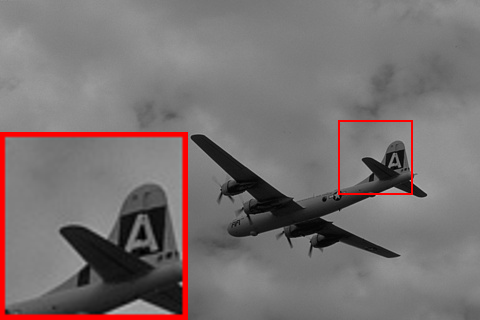}
{\scriptsize  Ground truth}
\end{minipage}
}
\caption{Examples of denoising results by different methods (noise level, $\sigma=50$).} 
\label{fig:denoising_visuals}
\end{figure}

\begin{table}[htp!]
\centering
\caption{Image Denoising PSNR results [dB] by different algorithms on BSD68.}
\label{tab:ID_results}
\resizebox{\linewidth}{!}
{
\begin{tabular}{c|cccc|c }
\hline\hline
 Noise Level& ~~BM3D~\cite{BM3D}~~& ~~WNNM~\cite{WNNM_CVPR}~~& ~~TNRD~\cite{TNRD}~~& ~~DnCNN~\cite{DnCNN}~~& ~~FDnet\\
\hline\hline          
 $\sigma=25$ &28.57&28.83&28.92&\color{red}29.23&\color{blue}29.12\\
  $\sigma=50$  &25.62&25.87&25.97&\color{blue}26.23&\color{red}26.24\\
   $\sigma=75$ &24.21&24.40&-&\color{blue}24.64&\color{red}24.76\\
\hline\hline
\end{tabular}
}
\end{table}

\section{Conclusion}
\label{sec:conclusion}

In this paper, we introduced the multi-bin trainable linear unit (MTLU), a novel activation function for increasing the nonlinear capacity of neural networks. 
MTLU was shown to be a robust alternative to the current activation functions, it improves the results of a wide range of restoration networks.
Based on MTLU, we proposed two efficient networks: a fast super-resolution network (FSRnet) and a fast denoising network (FDnet). 
FSRnet and FDnet are capable to better trade-off between speed and performance than prior art in image restoration.
On standard super-resolution and denoising benchmarks the proposed networks achieved comparable results with the current state-of-the-art deep learning networks but significantly faster and with lower memory requirements.

\section{Acknowledgement}
We gratefully acknowledge the support from NVIDIA Corporation
for providing GPU used in this
research.

{\small
\bibliographystyle{ieee}
\bibliography{egbib}
}

\end{document}